\documentclass[11pt]{article}
\usepackage{coling08}
\usepackage{times}
\usepackage{latexsym}
\usepackage{multirow}
\usepackage{epsf}
\usepackage{url}
\usepackage{amssymb}
\title{A Layered Grammar Model: Using Tree-Adjoining Grammars
to Build a Common Syntactic Kernel for Related Dialects}

\author{Pascal Vaillant\\
  GRIMAAG\\
  Universit{\'{e}} des Antilles et de la Guyane\\
  B.P. 792\\
  97337 Cayenne cedex\\
  (Guyane Fran{\c{c}}aise)\\
  {\tt pascal.vaillant@guyane.univ-ag.fr}}

\date{}

\begin{document}
\maketitle
\begin{abstract}
  This article describes the design of a common syntactic description
  for the core grammar of a group of related dialects. The common
  description does not rely on an abstract sub-linguistic structure
  like a metagrammar: it consists in a single FS-LTAG where the actual
  specific language is included as one of the attributes in the set of
  attribute types defined for the features. When the {\tt lan}
  attribute is instantiated, the selected subset of the grammar is
  equivalent to the grammar of one dialect. When it is not, we have a
  model of a hybrid multidialectal linguistic system. This principle
  is used for a group of creole languages of the West-Atlantic area,
  namely the French-based Creoles of Haiti, Guadeloupe, Martinique and
  French Guiana.
\end{abstract}

\section{Introduction}

Some of our present research aims at building formal linguistic
descriptions for regional languages of the area of the Lesser Antilles
and the Guianas, most of which are so-called ``under-resourced
languages''.  We have concentrated our efforts on a specific group of
languages, the French-based (or French-lexified) Creole languages of
the West-Atlantic area.  We are concerned with providing users of
those languages with electronic language resources, including formal
grammars fit to be used for various Natural Language Processing (NLP)
tasks, such as parsing or generation.

We are developing formal grammars in the TAG (Tree-Adjoining Grammars)
framework, the tree-centered unification-based syntactic formalism
which has proven successful in modelling other languages of different
types.  TAG grammars may be lexicalized, so they provide a
lexicon-centered description of phrase constructions
\cite{Schabes-Abeille-Joshi-1988}; and have been equipped with the
formal tool of double-plane feature structures, allowing the concept
of feature structures unification to get adapted to the specific needs
of adjunction~\cite{Vijay-Shanker-Joshi-1988}.

In the context we are working in, two practical reasons are leading to
the search of solutions for factoring as much as possible of the
grammars of those languages: first, the languages in this group are
fairly close to one another, with respect to both lexicon and grammar;
second, the resources dedicated to their description are scarce.  The
close relatedness makes it obvious for the linguist to try to leverage
the efforts spent on describing the grammar of one of the languages,
by factoring out all the common parts of the grammatical systems. This
principle has been used by other research work (see below,
Section~\ref{related-work}).

The originality of our approach is that we delay the point at which a
single language is actually chosen to the very last moment, namely at
generation time (the same would apply to parsing time, but parsing has
not been implemented yet). In the end, we propose a grammar which is
not a grammar for one single dialect, but a grammar for a
multidialectal complex, where language is one of the features selected
in the grammar itself, like person, number, tense, or aspect.

\section{Coverage of the grammar}
\label{coverage}

The portion of the grammar described so far represents only a small
fragment of the grammar of the languages we are interested in. Until
now, we have made attempts to describe: the determination of noun
phrases; the system of personal pronouns and determiners; the core
system of expression of tense, mood and aspect (TMA) of verbs~--- or, to
put it more cautiously, of predicates~---; the main auxiliary verbs
used to express other aspectual nuances; the expression of epistemic
and deontic modality; the combination of the negation with the above
mentioned subsystems (tense, aspect, modality) in the predicative
phrase.

The grammar and lexicon files are built upon an ad-hoc implementation
of FS-TAGs in Prolog\footnote{Precisely: SWI-Prolog, developed and
  maintained by Jan Wielemaker, University of Amsterdam:
  {\scriptsize\texttt{http://www.swi-prolog.org}}.}, which had
originally been developed in another context and for another language,
German~\cite{Vaillant-1999}, and later adapted to Martinican Creole
\cite{Vaillant-2003}.

The only function implemented at present is sentence generation; the
starting point of the generation is a conceptual graph, expressed by a
minimal set of spaning trees, which in turn select elementary trees in
the grammar (initial trees for the first pass, auxiliary trees for the
remaining parts). We are testing our grammar on a small sample tests
of such conceptual graphs.

In the remainder of this article, we will focus the attention on two
typical core subsystems of the grammar: determination in the noun
phrase, and expression of tense and aspect in the predicate
phrase\footnote{It may be inadequate to speak of {\em verb phrase} in
  the case of the Creole languages mentioned here, since any lexical
  unit (including nouns, but also some closed-class units like
  locative adverbs) may be inserted in the predicate slot of a
  sentence and bear tense or aspectual marks. So there probably are
  verbs, but possibly no ``verb phrases''~--- see \cite{Vaillant-2003}
  for a discussion.}.

\section{Application to French-based Creoles}

The family of dialects to which we apply the approach described is the
family of French-based (sometimes called French-lexified) Creole
languages of the West-Atlantic area.  Those languages emerged during
the peak period of the slave trade epoch (1650--1800) when France,
like some other West-European nations, founded colonies in the New
World and tried to develop intensive agricultural economic systems
based on the exploitation of slave workforce massively imported from
Africa.  In the quickly developing new societies, at any given moment
during that peak period, the number of people recently imported in any
colony tended to be higher than the number of people actually born
there~--- a typical situation for linguistic instability.  Moreover,
the slaves were brought from different regions of Africa and had no
common language to communicate with, except the language of the
European colons: so they were forced to use that target language,
without having time to learn it fully before passing it on to the next
generation of immigrants.  This type of situation leads to a very
specific drift of the language system, which begins to stabilize only
when the society itself stabilizes.  When observed in synchronicity at
the present moment, those Creoles obviously appear as languages which
share a very great portion of their vocabulary with French (more than
90\,\%), but have a very specific grammatical system, quite different
from the French one.

The languages falling into the category comprise French Creole
dialects born and developed in former French colonies of the Caribbean
Arc and its two continental ``pillars'': from the present US State of
Louisiana\footnote{A nearly extinct French Creole dialect~--- not to
  be confused with {\em Cajun} French~--- is still understood by some
  people in the parishes of Saint-Martin, Iberville and
  Pointe-Coup{\'{e}}e.} to French Guiana (formerly the Cayenne
colony), on the northern coast of the South-American
mainland. Caribbean islands where a French Creole has developed
include Hispaniola (in the western part of the island, the former
French colony of Saint-Domingue, since 1804 the independant republic
of Haiti), Guadeloupe, the island of Dominica, Martinique,
Saint-Lucia, and Trinidad (the latter also nearly extinct).  Among the
languages listed, we leave apart, for lack of easily accessible
sources and informants, the case of Louisiana, Dominica, Saint-Lucia
and Trinidad, and concentrate on the four Creoles of Haiti,
Guadeloupe, Martinique and French Guiana.

The question of how properly those languages qualify as a genetically
related family has been discussed in the literature.  A starting point
would be the obvious statement that all of them have French as an
ancestor\footnote{The atypical mode of language transmission has led
  some historical linguists~\cite[p.$\:$152]{Thomason-Kaufman-1988} to
  refuse to apply the term of genetic transmission, but this point has
  been thoroughly criticized~\cite{DeGraff-2005}.}, but this is not of
much linguistic interest since, as we have seen, the relatedness with
French lies principally in the vocabulary, whereas the Creole dialects
have a great convergence in their grammatical systems, that they
precisely do not owe to French.  Some formerly proposed theories of
monogenesis of {\em all} Creole languages are now largely out of
fashion; however, if the question is restricted to monogenesis of a
specific group of Creoles (e.g. French-based, or English-based) in a
specific region of the world (e.g. the West-Atlantic area),
monogenesis in this restricted acceptation remains a seriously
discussed hypothesis.  In any case, it has been established from
historical sources that there was uninterrupted contact and
interchange between the French colonies, from the first decades of
colonization up to now, so that it is a safe bet to consider the
different French Creole dialects as belonging to a dialect continuum.
Pf{\"{a}}nder~\shortcite[p.$\:$192--209]{Pfaender-2000}, notably,
proposes an analysis of the family in terms of dialectal area,
opposing center (Antilles) and periphery (Louisiana and Guiana), and
gives comparison tables for the systems of expression of tense and
aspect.

For a more detailed presentation of those languages, of their history,
and of the discussions they involve, the reader familiar with the
French language may easily access~\cite{Hazael-Massieux-2002}.

We will not enter into a detailed presentation of the grammatical
systems of the Creoles.  The most important thing to say here is that
they are isolating languages, SVO ordered, with a strict positional
syntax, and that tense and aspect are expressed by particles that are
placed before the main predicate.  As said above
(Section~\ref{coverage}), we will concentrate on the noun phrase and
on the TMA core system within the predicate phrase.  Tables \ref{np}
and \ref{predp} give an overview of those two systems.  They have been
compiled from different sources (most particularly
\cite{Pfaender-2000} and \cite{Damoiseau-2007} for the comparative
perspective, but also various other references for precise description
points specific to some given language), and completed following our
own observations on recent corpora.

\begin{table*}[hbtp]
\begin{center}
\begin{footnotesize}
\begin{tabular}{|l|l|l|l|l|l|l|}
\hline\hline
\multicolumn{2}{|l|}{\mbox{~}} & {\bf hait.} & {\bf guad.} & {\bf mart.} & {\bf guia.} & {\bf english} \\
\hline\hline
\multicolumn{2}{|l|}{Generic} & {\em moun} & {\em moun} & {\em moun} & {\em moun} & person (human) \\
\hline\hline
\multirow{7}{*}{Singular} & indefinite & {\em yon moun} & {\em on moun} & {\em an moun} & {\em roun moun} & a/one person \\
\cline{2-7}
 & \multirow{4}{*}{specific} & {\em moun nan} & {\em moun la} & {\em moun lan} & {\em moun an} & the person \\
\cline{3-7}
 & & {\em tab la} & {\em tab la} & {\em tab la} & {\em tab a} & the table \\
\cline{3-7}
 & & {\em chyen an} & {\em chyen la} & {\em chyen an} & {\em chyen an} & the dog \\
\cline{3-7}
 & & {\em zwazo a} & {\em zozyo la} & {\em zw{\'{e}}zo a} & {\em zozo a} & the bird \\
\cline{2-7}
 & \multirow{2}{*}{demonstrative} & {\em moun sa a} & {\em moun lasa} & {\em moun tala} & {\em sa moun an} & that person \\
\cline{3-7}
 & & {\em tab sa a} & {\em tab lasa} & {\em tab tala} & {\em sa tab a} & that table \\
\hline\hline
\multirow{7}{*}{Plural} & indefinite & {\em moun} & {\em moun} & {\em moun} & {\em moun} & people \\
\cline{2-7}
 & \multirow{4}{*}{specific} & {\em moun yo} & {\em s{\'{e}} moun la} & {\em s{\'{e}} moun lan} & {\em moun yan} & the persons \\
\cline{3-7}
 & & {\em tab yo} & {\em s{\'{e}} tab la} & {\em s{\'{e}} tab la} & {\em tab ya} & the tables \\
\cline{3-7}
 & & {\em chyen yo} & {\em s{\'{e}} chyen la} & {\em s{\'{e}} chyen an} & {\em chyen yan} & the dogs \\
\cline{3-7}
 & & {\em zwazo yo} & {\em s{\'{e}} zozyo la} & {\em s{\'{e}} zw{\'{e}}zo a} & {\em zozo ya} & the birds \\
\cline{2-7}
 & \multirow{2}{*}{demonstrative} & {\em moun sa yo} & {\em s{\'{e}} moun lasa} & {\em s{\'{e}} moun tala} & {\em sa moun yan} & those people \\
\cline{3-7}
 & & {\em tab sa yo} & {\em s{\'{e}} tab lasa} & {\em s{\'{e}} tab tala} & {\em sa tab ya} & those tables \\
\hline\hline
\end{tabular}
\end{footnotesize}
\end{center}
\caption[Noun Phrase]{Determination in the noun phrase}
\label{np}
\end{table*}

\subsection{Determination in the noun phrase}

The four Creoles all possess four systematic degrees of determination
of nouns: a generic, an indefinite, a specific, and a demonstrative.
The generic is used when the concept is taken for its general features
as a category; in English, the same meaning could sometimes be
expressed with a singular, and sometimes with a plural ({\em zwazo gen
  de zel} (hait.): the bird has two wings$\:$/$\:$a bird has two
wings$\:$/$\:$birds have two wings).  For the sake of descriptive
economy, in the formalization, we treat this generic degree as simply
being one of the possible semantic values of the plural indefinite
(which is also expressed by the bare noun, with no
article)\footnote{This interpretation agrees with a number of
  linguistic facts, like anaphora {\em often} involving a plural
  pronoun ({\em zwazo gen de zel pou yo kapab vole}: bird$[$s$]$ have
  two wing$[$s$]$ for them $[$to be$]$ able $[$to$]$ fly).}.  The
indefinite degree, like in French or German, is expressed by a numeral
(and its value is more specific, closer to the original semantics of
the numeral, than it has become in French, for instance~--- where the
indefinite article also is used to express the generic).  The specific
degree (roughly equivalent to English ``the'') is expressed by a
postposed article, historically deriving from a French deictic adverb
({\em l{\`{a}}}).  Lastly, the demonstrative degree derives from the
combination of a former demonstrative pronoun, now sometimes preposed
(guia.) and sometimes postposed (other Creoles) to the noun, and to
which the mark of the specific definite is added (with a case of fused
form for Guadeloupean and Martinican).

The plural is expressed either by a preposed marker derived from a
former plural demonstrative (mart., guad.), or by a postposed
third-person plural personal pronoun (hait., guia.), which in the case
of guianese got fused with the definite mark ({\em y{\'{e}} la}
$[$historical form, described in 1872$]$ $>$ {\em ya} $[$contemporary
  form$]$\/).

In our formal model, we only keep three degrees of determination
(indefinite, specific and demonstrative), which combine with two
values for number (singular and plural).  Also, since the indefinite
mark does not combine with the others (when in contrast, there is a
combination between the marks of demonstrative and specific, with
demonstrative $\Rightarrow{}$ specific), we model the indefinite by an
absence of determination feature; the specific is modeled by the
feature $\langle{}\mbox{spe}=+\rangle{}$\/; and the demonstrative by
the combination of features
${\;\langle{}\mbox{spe}=+\rangle{},\,\langle{}\mbox{dem}=+\rangle{}\;}$\/.

In some dialects, a phenomenon of nasal progressive assimilation
changes the surface form of the postposed specific article (hait.,
mart., guia.); in others, in addition, the surface form of the article
differs depending on whether the preceding word ends with a vowel or a
consonant (hait., mart.).  The four possible combinations are shown in
table~\ref{np}.

\begin{table*}[hbtp]
\begin{center}
\begin{footnotesize}
\begin{tabular}{|l|l|l|l|l|}
\hline\hline
\mbox{~} & {\bf hait.} & {\bf guad.} & {\bf mart.} & {\bf guia.} \\
\hline\hline
Accomplished\,/\,Aoristic & {\em danse} & {\em dans{\'{e}}} & {\em dans{\'{e}}} & {\em dans{\'{e}}} \\
Unaccomplished\,/\,Present & {\em danse} & {\em ka dans{\'{e}}} & {\em ka dans{\'{e}}} & {\em (ka) dans{\'{e}}} \\
Frequentative & {\em danse} & {\em ka dans{\'{e}}} & {\em ka dans{\'{e}}} & {\em ka dans{\'{e}}} \\
Progressive & {\em ap danse} & {\em ka dans{\'{e}}} & {\em ka dans{\'{e}}} & {\em ka dans{\'{e}}} \\
Near Future & {\em pral danse} & {\em kay dans{\'{e}}} & {\em kay dans{\'{e}}} & {\em k'al{\'{e}}\,/\,kay dans{\'{e}}}\\
Future & {\em va danse} & {\em k{\'{e}} dans{\'{e}}} & {\em k{\'{e}} dans{\'{e}}} & {\em k{\'{e}} dans{\'{e}}} \\
Unaccomplished Future {\em (seldom)} & {\em vap danse} & {\em k{\'{e}} ka dans{\'{e}}} & {\em k{\'{e}} ka dans{\'{e}}} & {\em k{\'{e}} ka dans{\'{e}}} \\
Accomplished past (pluperfect) & {\em te danse} & {\em t{\'{e}} dans{\'{e}}} & {\em t{\'{e}} dans{\'{e}}} & {\em t{\'{e}} dans{\'{e}}} \\
Unaccomplished past & {\em tap danse} & {\em t{\'{e}} ka dans{\'{e}}} & {\em t{\'{e}} ka dans{\'{e}}} & {\em t{\'{e}} ka dans{\'{e}}} \\
Irrealis & {\em ta danse} & {\em t{\'{e}} k{\'{e}} dans{\'{e}}} & {\em t{\'{e}} k{\'{e}} dans{\'{e}}} & {\em t{\'{e}} k{\'{e}} dans{\'{e}}} \\
Irrealis unaccomplished & {\em ta vap danse} & {\em t{\'{e}} k{\'{e}} ka dans{\'{e}}} & {\em t{\'{e}} k{\'{e}} ka dans{\'{e}}} & {\em t{\'{e}} k{\'{e}} ka dans{\'{e}}} \\
Conditional\,/\,Optative & {\em ta danse} & {\em t{\'{e}} k{\'{e}} dans{\'{e}}} & {\em s{\'{e}} dans{\'{e}}} & {\em t{\'{e}} k{\'{e}} dans{\'{e}}} \\
\hline\hline
\end{tabular}
\end{footnotesize}
\caption[Predicative phrase]{Core tense and aspect marking in the predicative phrase}
\label{predp}
\end{center}
\end{table*}

\subsection{Tense and aspect in the predicative phrase}

In Creole linguistics, a classical description given of the TMA
(Tense-Mood-Aspect) system of the ``Atlantic'' Creole
languages\footnote{The schema also holds for English-based
  Creoles~\cite{Bickerton-1981}.} mentions three optional components
appearing in a very strict order: past tense mark; ``mood'' mark (able
to take future or irrealis values, depending on contexts);
imperfective aspect mark.  A canonical version of this system has been
given for French-based Creoles by Valdman~\shortcite{Valdman-1978},
who actually describes those three categories as one category of tense
({\em past}\,) and two categories of aspect ({\em prospective} and
{\em continuative}\/).  The ``middle'' mark (Valdman's
``prospective'') takes on an irrealis meaning when it is combined with
the past tense.

So, there is a combinatory system: ({\em
  t{\'{e}}}\,/\,$\varnothing{}$) $\times{}$ ({\em
  k{\'{e}}}\,/\,$\varnothing{}$) $\times{}$ ({\em
  ka}\,/\,$\varnothing{}$) (if we call the three marks by the form
they have in the three Creoles of Guadeloupe, Martinique and Guyane),
which in theory generates eight possible combinations: {\em
  $\varnothing{}$}, {\em ka}, {\em k{\'{e}}}, {\em k{\'{e}} ka}, {\em
  t{\'{e}}}, {\em t{\'{e}} ka}, {\em t{\'{e}} k{\'{e}}}, {\em t{\'{e}}
  k{\'{e}} ka}.  The eight combinations are attested to different
degrees, with the semantic values given in table~\ref{predp}.  In
Haitian Creole, the corresponding forms are {\em te}, {\em va} and
{\em ap}, and some combinations yield fused forms (va ap $>$ vap; te
ap $>$ tap; te va $>$ ta; te va ap $>$ ta vap).

In fact, there are variations in this basic schema. For instance, the
term ``imperfective'' covers a complex of diverse meanings
(progressive, frequentative, or simply unaccomplished) which do not
strictly overlap in the different dialects.  For instance, if the mark
{\em ka} may bear all the above-mentioned meanings in the Creoles of
Guadeloupe or Martinique (up to some general temporal value roughly
corresponding to the English simple present), it is not necessarily so
in the Creole of Guiana, and it is quite false for the Creole of Haiti
(where the unaccomplished is unmarked, and the only aspectual value of
particle {\em ap} is the progressive, corresponding not to English
simple present, but to English {\sc be} + {\em -ing}~--- and even able
to take over the temporal value of a future). Table~\ref{predp} shows
these differences.

Lastly, it is important to notice that the combinations of the TMA
marks are constrained by the semantics of the unit placed in the
predicate position.  For instance, a verb with a ``non-processual''
meaning (like {\em kon{\`{e}}t}, to know), or an adjective referring
to a state (like {\em malad}, ill), will hardly combine with an
imperfective aspect marker like {\em ka}; if they do, however, it will
necessarily produce a meaning effect that will shift the contextual
meaning towards a less ``stative'' value.  For example, an utterance
like {\em mo ka malad} (I-{\sc imp}-ill) might be attested; and it is
to be interpreted, depending on the context, either as a frequentative
(at every back to school time, I get flu), or as a progressive (I feel
I am coming down to flu).

\subsection{Some TAG model elements}

\begin{figure}[hbtp]
\begin{center}
\epsfxsize=60mm
\mbox{\epsffile{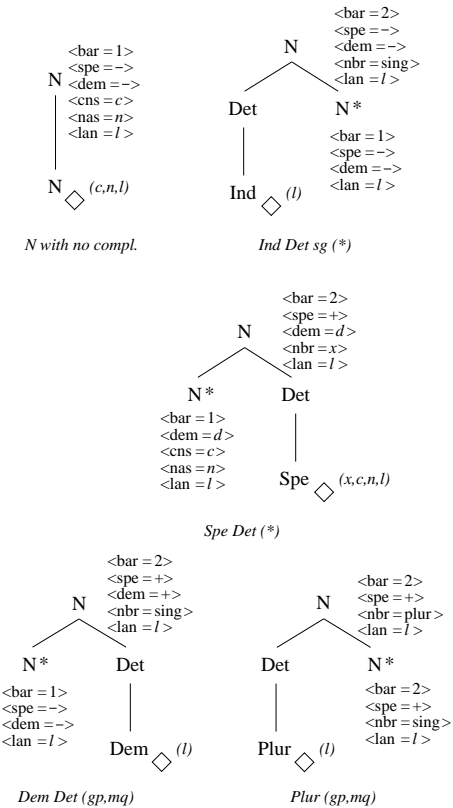}}
\caption[NP~--- Common elements]{Common elements in the NP model.}
\label{np-comm}
\end{center}
\end{figure}

\begin{figure}[hbtp]
\begin{center}
\epsfxsize=60mm
\mbox{\epsffile{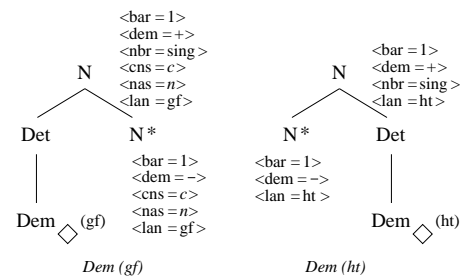}}
\caption[NP~--- Specific elements]{NP modelling elements specific
  to haitian and guianese}
\label{np-spec}
\end{center}
\end{figure}

In figures~\ref{np-comm} and \ref{np-spec}, we show the main
components of the model for the noun phrase system presented in
table~\ref{np}, represented as elementary trees with a language
parameter $l$\,\footnote{The following abbreviations are used for the
  attributes: {\em bar} = bar level (1 = noun with complements, but no
  determination; 2 = noun phrase); {\em nbr} = number; {\em spe} =
  specific determiner; {\em dem} = demonstrative determiner; {\em cns}
  = the constituent ends with a consonant; {\em nas} = the constituent
  ends with a nasal syllable; {\em lan} = language. The values used to
  identify the four Creoles are based on the two-letter country codes
  defined in standard ISO-3166 for country names: HT for {\em
    Haiti}\/, GP for {\em Guadeloupe}\/, MQ for {\em Martinique}\/,
  and GF for {\em French Guiana} (going from North to South... and by
  decreasing population count.)  Non-instantiated variables are in
  italics.}.

It should be noted that the trees {\em Dem Det (gp,mq)} and {\em Plur
  (gp,mq)}, which concern only two dialects among the four
(Guadeloupean and Martinican), are included in the common layer
without risking to interfere with the construction of the
demonstrative or plural in Haitian or Guianese (in fact, unification
constraints forbid the adjunction of a GP/MQ demonstrative on a HT/GF
demonstrative; likewise, they forbid the adjunction of a GP/MQ plural
on a HT/GF plural).

The adjunction of the demonstrative in Haitian or Guianese is done
above the level of the noun complements (attention to parameter {\em
  bar} in the trees {\em Dem (gf)} et {\em Dem (ht)}\/), but below the
specific article; e.g.~{\em moun Sentoma sa yo} (hait.): those people
from Saint-Thomas; {\em sa moun Senloran an} (guia.): those people
from Saint-Laurent.

\begin{figure}[hbtp]
\begin{center}
\epsfxsize=60mm
\mbox{\epsffile{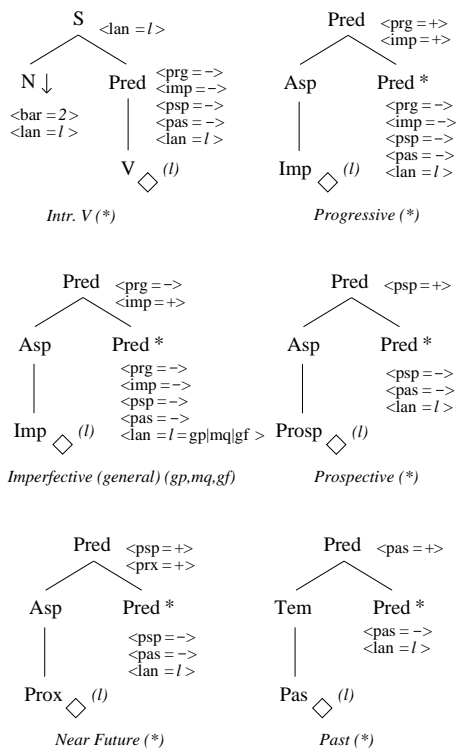}}
\caption[Pred P~--- Common elements]{Common elements in the predicative phrase model}
\label{predp-comm}
\end{center}
\end{figure}

The TMA system, on its side, is in a great part common to the four
languages.  Auxiliary trees modelling the adjunction of aspectual or
temporal values hence are all common (fig.~\ref{predp-comm}). The only
nuance resides in the fact that the tree for adjoining an aspect
particle to convey general values of imperfective (durative,
frequentative) cannot unify when the {\tt lan} parameter is set to
Haitian.  In the end, only the lexical (surface) values make the
differences between the dialects\footnote{The following abbreviations
  are used for the attributes in fig.~\ref{predp-comm}: Tense: {\em
    pas} = past; Aspects: {\em psp} = prospective; {\em prx} =
  proximal prospective (``imminent'' aspect~\textasciitilde{} temporal
  value of a near future)~; {\em imp} = imperfective (general)~; {\em
    prg} = progressive (like in English ``I am doing...'').}.

\section{Related work}
\label{related-work}

The idea of factoring some of the efforts of grammar modelling to
exploit similar structures among different languages has already been
tackled by some research works, among which we are particularly aware
of those led at Jussieu within the FTAG project~\cite{Candito-1998},
the Lexorg project~\cite{Xia-et-al-1998}, the LinGO grammar
Matrix~\cite{Bender-et-al-2002,Bender-Flickinger-2005}\footnote{See
  LinGO grammar matrix' web site:
  {\scriptsize\texttt{http://www.delph-in.net/matrix/}}.}, the LORIA
XMG project~\cite{Crabbe-et-al-2006}, and Bouillon et al.'s
work~\shortcite{Bouillon-et-al-2006} on multilingual multipurpose
grammars\footnote{Thanks to the reviewers of the preliminary version
  of this article for pointing to some useful references.}.

Works like Candito's~\shortcite{Candito-1998} (for French and Italian)
or Xia and Palmer's~\shortcite{Xia-et-al-1998} (for English and
Chinese), are based on the idea of using {\em metagrammars}, that is
higher-level descriptions of general properties of the language(s)
described.  The higher-level descriptions for different languages may
be factored as long as the languages share typological features.  In
the end, an actual LTAG grammar is generated from the meta-grammar,
tailored for one specific language.  In this type of approaches, what
is actually shared between the languages is a higher-level structure,
not actual grammatical structures belonging to the LTAG description of
the languages.

In the LinGO grammar matrix approach~\cite{Bender-et-al-2002},
underspecified HPSG structures (with a minimal recursion semantics)
are used to share information between different languages.  A system
based on shell scripts is used to automatically generate grammar files
for a specific language, when given a couple of general typological
specific information (word order pattern, case marking strategy,
etc.).

The approach which most resembles the one advocated in the present
paper is Bouillon et al.'s~\shortcite{Bouillon-et-al-2006} way of
devising quickly re-usable grammars for speech recognition programs,
based on shared grammatical descriptions for related romance languages
(French, Castilian Spanish, and Catalan).  The authors include
``macros'' in their DCG-style upper-level description, and the macros
allow to specify alternative points where the languages differ (like
the position of clitics in specific verb forms, the optionality of
determiners, the optional presence of prepositions for object
complements, etc.). In a last stage, the DCG-style specification is
compiled to ad hoc CFGs tailored for speech recognition engines, each
for a specific language and task.

Our approach, in contrast, is not a meta-grammar approach; what is
shared between the different languages are actual LTAG trees.  The
``language'' parameter is embedded in the very feature structures of
tree nodes.  So, our lexical-grammatical descriptions reside in one
single level of description, but that level is ``modularized'': some
descriptions are common to all the dialects described, some are shared
by only part of them, and some are specific.  In other works, even
those which are not based on meta-grammars (like Bender's or
Bouillon's), the goal is to generate a grammar for a single language
in the end.  In the present work we are aiming at giving a description
of a multidialectal linguistic system.

\section{Discussion}
\label{discussion}

The above-mentioned modelling choice may seem counter-intuitive in the
theoretical frame of structural linguistics.  One might object that if
the language itself is the whole object of description, then it is
absurd to include it as a category in the description.  This view is
justified as long as one does not wish to take into account dialectal
variation as an internal system variable.  If this is the case, then
every single dialect must be considered an isolate and be given a
holistic, unitary description.

But in the context we are working in, several rationales lead us to
think that it might be a good idea to include dialectal variation in
the description.

We already have mentioned practical reasons (see above, in
Introduction).  The ``time saving'' and ``resource sharing''
rationales applies to our method as well as to others (like
meta-grammars).  A supplementary argument, which applies more
specifically to our method, is the fact that in the cases we are
studying, not only some syntactical properties of the languages are
common, but also an important part of the vocabulary, until at the
very surface level.  This speaks for sharing bottom-level structures.

But there is another, less practical, type of argument: if we have a
modular grammatical system model which ``contains'' more than one
language in itself, we are able to model the linguistic competence in
one of the languages, but also to model multilingual (in the present
case, multidialectal) linguistic competence.

If our goal is to model monolingual competence, this is easily done by
unifying the {\tt lan} parameter with one of its possible values, and
then erasing the (now redundant) parameter from the description.

However, in some cases, we might want to have a model of
multidialectal variation.  Considered from the {E}-language side, we
then have a model of a dialectal continuum.  Considered from the
{I}-language side, we have a model of the linguistic competence of a
multilingual speaker of related dialects.  The interplay of
grammatical structures of a multidialectal system, the possibilities
of combination and unification given different levels of instantiation
of the {\tt lan} parameter, might provide us with a model for such
linguistic phenomena as: specialized repertoires, code switching, code
mixing, or {\em koin{\^{e}}} emergence.  That work, at the present
stage, is still to be done: it is a mere idea of future research
directions to evaluate the potential of our modelling method.  Yet it
is an appealing idea, given that in some types of contexts,
multilinguality among related dialects is a common
situation\footnote{This is particularly the case in the regions where
some of the languages we study are spoken: for example, in Guadeloupe
and in French Guiana, there are communities of tens of thousands of
people of Haitian descent, who tend to mix the Creole of Haiti with
the Creole of the country.  In the European mainland part of France,
there also are large numbers of people from the French West Indies,
who tend to form multidialectal speakers communities, where specific
Creole differences between e.g. Guadeloupe and Martinique are
vanishing.}, and that phenomena such as code switching or code mixing
are more frequent than the opposite~--- the use of a single unitary
language with a single norm\footnote{In another study, presented
elsewhere~\cite{Lengrai-Moustin-Vaillant-2006}, we have shown that
within a corpus of several hours of recorded radio broadcastings in
Creole of the Martinique, it is hard to find a single minute of speech
where French and Creole are not mixed at the very intra-sentential
level.}.  It is also a matter of future research to evaluate the
degree of parsing feasibility for mixed linguistic input.

\bibliographystyle{coling}
\bibliography{biblio-layered-grammar}

\end{document}